# Text Production and Comprehension by Human and Artificial Intelligence: Interdisciplinary Workshop Report

**Participants:**

Patrick Bolger, Nottingham Trent University

Morten H. Christiansen, Cornell University

Bill Cope, University of Illinois Urbana-Champaign

Pablo Contreras Kallens, University of the Saarland

Mary Kalantzis, University of Illinois Urbana-Champaign

Kyle Mahowald, University of Texas at Austin

Martin Pickering, University of Edinburgh

Michael Ramscar, Tuebingen University

Alina Reznitskaya, Montclair State University

Leila Wehbe, Carnegie Mellon University

Wei Xu, Georgia Institute of Technology

**Organizers:**

Evgeny Chukharev, Iowa State University

Emily Dux Speltz, Embry-Riddle Aeronautical University

Mark Torrance, Nottingham Trent University

**Discussants:**

Kimberly Becker, Midland University

Phuong Nguyen, University of Chicago

Dan Song, University of Iowa

**Report Author:**

Emily Dux Speltz, duxspele@erau.edu

This material is based upon work supported by the National Science Foundation Programs in Science of Learning and Augmented Intelligence, Linguistics, and Robust Intelligence under Grant No. 2422404. Any opinions, findings, and conclusions or recommendations expressed in this material are those of the authors and do not necessarily reflect the views of the National Science Foundation.

# Table of Contents

# Executive Summary

This report synthesizes the outcomes of a recent interdisciplinary workshop that brought together leading experts in cognitive psychology, language learning, and artificial intelligence (AI)-based natural language processing (NLP)[1]. The workshop, funded by the National Science Foundation, aimed to address a critical knowledge gap in our understanding of the relationship between AI language models and human cognitive processes in text comprehension and composition.

Through collaborative dialogue across cognitive, linguistic, and technological perspectives, workshop participants examined the underlying processes involved when humans produce and comprehend text, and how AI can both inform our understanding of these processes and augment human capabilities. The workshop revealed emerging patterns in the relationship between large language models (LLMs)[2] and human cognition, with highlights on both the capabilities of LLMs and their limitations in fully replicating human-like language understanding and generation.

Key findings include the potential of LLMs to offer insights into human language processing, the increasing alignment between LLM behavior and human language processing when models are fine-tuned with human feedback, and the opportunities and challenges presented by human-AI collaboration in language tasks.

This report outlines several recommendations to address the identified knowledge gap:

1. Developing interdisciplinary research programs to create integrated models of human-AI language processing.

2. Designing targeted studies to examine human-AI interaction in collaborative text tasks.

---

[1] Natural language processing (NLP) is a field at the intersection of artificial intelligence, linguistics, computer science, logic, and psychology (Joshi, 1991; Nadkarni et al., 2011) in which various aspects of language are studied through mathematical and computational modeling (Reshamwala et al., 2013).

[2] Large language models (LLMs) are AI systems trained on vast amounts of text data to generate human-like language (Teubner et al., 2023).

3. Initiating long-term studies to assess the impact of sustained human-AI collaboration on language skills and cognition.
4. Creating educational interventions that leverage AI-human collaboration while addressing potential risks.
5. Establishing ethical guidelines for AI development and implementation in language-related tasks.

By synthesizing these findings and recommendations, this report aims to guide future research, development, and implementation of LLMs in cognitive psychology, linguistics, and education. It emphasizes the importance of ethical considerations and responsible use of AI technologies while striving to enhance human capabilities in text comprehension and production through effective human-AI collaboration.

# 1. Introduction

This report explores the state of the field regarding text production and comprehension by human and artificial intelligence by summarizing insights from a recent workshop that brought together leading researchers. It examines the impact of LLMs on creativity, the role of human feedback in refining AI performance, theoretical and practical applications in education, and the cognitive processes underlying language learning.

The workshop, "Text Production and Comprehension by Human and Artificial Intelligence," was convened to explore the cognitive and technological processes underlying the comprehension and composition of cohesive multi-sentence texts. By bringing together leading researchers in cognitive psychology, linguistics, and AI-based NLP, the workshop aimed to address two interrelated questions:

1. To what extent do LLMs offer hypotheses about how human minds produce natural-language texts?
2. What cognitive processes occur in human writers' minds when they collaborate with AI systems, such as ChatGPT, in text production?

# 2. Background

Historically, the development of natural language generation (NLG) and natural language understanding (NLU) systems, on the one hand, and the research into cognitive processes that underlie written text composition and comprehension in humans, on the other, have followed independent but complementary trajectories. For instance, early NLG systems (e.g., PAULINE) were based on symbolic models driven by explicit communicative goals expressed as state-action rules (Hovy, 1990). Cognitive-psychological research in text composition was also dominated by the assumption that successful writers plan their texts explicitly through a conscious and deliberate selection of actions to achieve goals (Alamargot et al., 2011; Hayes & Flower, 1986; Kellogg, 1999). This research was rooted in theory around general problem solving (e.g., Alamargot & Chanquoy, 2012), both in humans and computers.

Increasingly, however, problem-solving accounts of human text planning appear untenable. Studies that precisely determine the time course of written production suggest timescales that are often too brief to allow for the possibility of explicit rule-based decision-making: Coherent, intelligent text can be produced without the writer stopping to think (Chukharev-Hudilainen et al., 2019). It seems at least plausible, therefore, that the logical coherence structure of completed texts is not primarily driven by a logical, coherent representation in the writer's mind. The behaviors that writers engage in (such as glances back in the existing text) may instead enact implicit, low-level, probabilistic, and associative processing. This understanding of text production may, in important ways, parallel what happens within NLG systems like ChatGPT. Therefore, there is considerable potential to explore the potential of AI-based NLG models as a source of hypotheses for the processing that occurs in human text composition. The same would apply to text comprehension, both as an independent activity and as part of the source-based composition process.

Further, LLMs have substantial potential for text understanding and generation not only in the absence of humans, but (probably more importantly) collaboratively with humans. ChatGPT, for example, affords dialogic interaction with a partially unreliable source that rapidly generates focused, well-formed text (e.g., Labruna et al., 2023; OpenAI, 2025; Waldo & Boussard, 2024). This is quite different from solo writing from textual sources typical in most adult written composition contexts. Instead, it is much more akin to synchronous collaboration with a human co-author. There are, however, important differences: ChatGPT knows more (with caveats) and produces text more quickly than a human. There are obviously many questions around how this interaction is most effectively managed, both to generate coherent and effective text, and to maximize learning in the human reader/writer. A more fundamental question, however, is how interaction with ChatGPT meshes with the (arguably) implicit and finely balanced processing that results in text comprehension and generation in humans. How does a human writer's mind accommodate interaction with and input from ChatGPT in real time when gaining an understanding of and producing texts?

We contend that there is a need for an in-depth exploration of the relationship between AI language models and cognitive psychology, specifically focusing on the processes of text comprehension and composition. By bringing together experts in cognitive psychology, linguistics, and AI-based NLP, our workshop aimed to address fundamental questions regarding the representation of written language knowledge in humans, the cognitive processes involved in human-AI collaboration during text comprehension and composition, and the potential of LLMs for enhancing reading and writing skills. This interdisciplinary dialogue is crucial for unlocking new insights, fostering innovative learning experiences, and optimizing the interaction between AI and human cognition in the context of written text processing.

### 3. Problem Statement

The rapid advancement of AI language models has created a gap in our understanding of their interaction with human cognitive processes, particularly in the domains of text comprehension and composition. This knowledge deficit poses a substantial challenge to effectively leveraging AI technologies in educational and professional settings. The absence of a comprehensive framework integrating insights from cognitive psychology, linguistics, and AI-based NLP could be hindering our ability to optimize human-AI collaboration and impeding the enhancement of reading and writing skills across various demographics.

The complexity of this challenge is further amplified by the potential dual impact of LLMs on literacy, as highlighted by recent research. On one hand, there are concerns that over-reliance on LLMs could lead to decreased perceptual and cognitive skills, poorer domain- and culturally-specific knowledge, increased susceptibility to misinformation, and potentially accelerated IQ decline. There are also worries about the amplification of stereotypes and other issues in LLMs' training data (Bender et al., 2021). Conversely, LLMs present significant opportunities to enhance literacy (Huettig & Christiansen, 2024). When used critically, researchers have argued that they could facilitate the writing process (Adams & Chuah, 2022), level the playing field for disadvantaged writers (Berdejo-Espinola & Amano, 2023), aid in synthesizing vast literatures (Agathokleous et al., 2024), and potentially promote literacy by incorporating turn-taking into reading (Levinson, 2016).

The root cause of this lack of a comprehensive framework lies in the traditionally siloed nature of these disciplines, which has prevented the cross-pollination of ideas necessary for a holistic understanding of AI-human cognitive interactions. As AI language models become increasingly sophisticated and ubiquitous, the lack of interdisciplinary research in this area may lead to suboptimal educational practices, missed opportunities for cognitive enhancement, and potential misalignments between AI systems and human cognitive processes.

The consequences of failing to address this knowledge gap could be far-reaching. In educational contexts, we risk developing pedagogical approaches that do not fully capitalize on the potential synergies between AI and human cognition, potentially stunting students' intellectual growth and adaptability in an AI-driven world. In professional environments, the absence of a nuanced understanding of AI-human cognitive interactions may result in decreased productivity, increased cognitive load, and missed opportunities for innovation. Moreover, without a deep understanding of how AI language models interface with human cognitive processes, we may inadvertently exacerbate existing inequalities in access to education and information. This could lead to a widening skills gap in society, with those able to effectively leverage AI technologies gaining a significant advantage over those who cannot.

To address these pressing concerns, our workshop facilitated an interdisciplinary dialogue among experts in cognitive psychology, linguistics, and AI-based NLP. This collaborative effort aimed to bridge the existing knowledge gap and enhance our understanding of the cognitive processes involved in human-AI collaboration. The insights gained from this exploration are expected to inform the development of more effective educational practices, optimize the interaction between AI and human cognition, and potentially mitigate the risks associated with the unchecked proliferation of AI language models in various spheres of human activity.

## 4. Methodology

The workshop was designed to foster collaboration and dialogue across these disciplines. The primary objective was to explore the intersections of human cognition and artificial intelligence in language processing, with a particular focus on the following subtopics:

- Statistical learning of language
- Computational modeling of language acquisition
- Language production and comprehension
- Collaborative learning
- LLMs
- Human-and-computer-in-the-loop learning
- Text mining and machine learning
- Explainable AI in education
- Human-computer interaction (HCI)

### 4.1 Workshop Structure

Based on the proposals submitted by invited speakers, the discussions were organized around three key topics:

1. The potential of large language models to offer insights into human language processing.
2. The cognitive processes involved in human-LLM collaboration on language tasks.
3. The effectiveness of LLM-based tools in enhancing learning in language-intensive domains.

### 4.2 Data Collection and Analysis

The insights presented in this report are derived from presentations by invited speakers, panel discussions, and Q&A sessions. Each presentation and discussion was carefully documented and analyzed to extract key themes, challenges, and opportunities for future research and collaboration.

The following sections present a comprehensive summary and synthesis of the workshop presentations, organized by the three main topics. This structure allows for a systematic exploration of the current state of research, emerging trends, and potential future directions in the intersection of cognitive psychology and AI-based NLP.

By facilitating this unique interdisciplinary dialogue, the workshop has laid the groundwork for future collaborations and research initiatives that could significantly advance our understanding of both human and artificial language processing.

## 5. Findings

### 5.1 Topic 1: To what extent do large language models offer hypotheses about how human minds/brains produce (or comprehend) natural-language texts?

Recent advancements in LLMs have sparked considerable interest in their potential to inform our understanding of human language processing. This section synthesizes insights from three leading researchers who presented their work on the intersection of artificial intelligence and human cognition in language production and comprehension.

**Wei Xu**'s research focuses on practical applications of Natural Language Processing (NLP), addressing key challenges in text simplification (Heineman et al., 2023), privacy preservation (Dou et al., 2024), and mitigation of cultural biases (Naous et al., 2024). While evaluating GPT-4 for text simplification, Xu's team found that the model demonstrates high fluency but still requires human oversight to ensure accuracy. This finding underscores both the capabilities and limitations of current AI models in mimicking human-like language understanding and generation. Xu's work on privacy protection tools and culturally sensitive NLP highlights the importance of interdisciplinary collaboration in enhancing the accessibility, privacy, and cultural sensitivity of AI-driven language processing.

**Leila Wehbe**'s presentation explored innovative approaches to using language models for studying the neurobiology of language, emphasizing the need for flexibility beyond simple alignments of neural networks and brain activity. Wehbe posed fundamental questions about breaking down real-life language production into components and addressing more complex tasks. Her research shows the interplay between stimuli, brain activity, and neural networks in language processing. Wehbe highlighted research on fine-tuning language models to predict fMRI and MEG data, demonstrating improved generalizability across subjects and imaging modalities, though without significant improvements in NLP task performance. The talk also addressed modeling active language tasks, such as question answering, by developing methods to capture representations of both task and stimulus interactions. Wehbe discussed the potential for extending these models to more interactive setups and emphasized the importance of integrating large language models' conversational abilities to study the brain basis of dialogue. Additionally, she underscored the need to bridge the gap between language comprehension and production research, particularly in semantic representation studies, to advance our understanding of language processing in the brain.

**Michael Ramscar**'s presentation challenged traditional assumptions about language processing, proposing a discriminative learning approach based on uncertainty reduction rather than compositional meaning transfer. Ramscar's framework aligns more closely with how LLMs learn, suggesting that these models may offer valuable insights into human language learning and processing. His research on the statistical structure of language, the bursty nature of word occurrences (where words tend to appear multiple times in specific contexts rather than being evenly distributed), and the role of grammatical systems in managing noun unpredictability provides a theoretical bridge between human cognition and machine learning approaches. This perspective indicates that LLMs, while not perfect analogies for human minds, can offer hypotheses about certain aspects of human language processing, particularly in terms of learning mechanisms and the management of linguistic uncertainty.

Collectively, these presentations offer diverse perspectives on how LLMs can inform our understanding of human language processing. They demonstrate the potential of using AI models as tools for studying the neurobiology of language and generating hypotheses about human cognition. However,

Page | 10

the research also highlights the limitations of current models in fully capturing the nuance and contextual understanding characteristic of human language use.

The findings presented suggest that while LLMs can provide valuable insights into certain aspects of language processing, they should not be considered direct replicas of human cognitive processes. Instead, they offer a complementary approach to studying language, potentially revealing patterns and mechanisms that may be difficult to observe through traditional methods alone. As the field progresses, it is clear that interdisciplinary collaboration will be crucial. Combining insights from cognitive psychology, neuroscience, linguistics, and computer science will be essential for developing more comprehensive models of human language processing and for creating AI systems that can better emulate human-like language understanding and generation.

In summary, while LLMs offer promising avenues for generating hypotheses about human language processing, researchers should remain cautious in drawing direct parallels between AI and human cognition. The complex nature of human language use, influenced by factors such as culture, emotion, and individual experience, necessitates a nuanced approach to integrating insights from AI models into our understanding of human language production and comprehension.

**5.2 Topic 2: What might happen in humans' minds as they dialogue/collaborate with large language models to perform language tasks?**

The rapid advancement of LLMs has opened new opportunities for human-AI collaboration, particularly in the realm of language tasks. As these AI systems become increasingly sophisticated and integrated into various aspects of communication and writing, it is valuable to understand the cognitive implications of such interactions. This section explores the potential effects on human cognition when individuals engage with LLMs in language-related activities. Drawing from the presentations of **Martin Pickering, Kyle Mahowald, Pablo Contreras Kallens,** and **Morten H. Christiansen**, we examine the complex interplay between human mental processes and AI assistance, considering both the opportunities and challenges that arise from this collaboration.

**Martin Pickering**'s presentation focused on the concept of augmented language production, exploring how humans might interact with AI assistants during speech and writing tasks. Pickering proposed that successful collaboration between humans and LLMs depends on an alignment of representations, similar to human-to-human dialogue. He suggested that LLMs could assist in real-time speech production by prompting forgotten words, guiding topic selection, and correcting mistakes, as illustrated in Figure 1. This diagram outlines various ways an AI system could augment human speech production, from providing relevant contextual information to assisting with formulation and comprehension. For written composition, Pickering emphasized the importance of LLMs developing similar representations to humans for effective collaboration, particularly in understanding context and intended meanings.

**Figure 1**.

Pickering's diagram illustrating how an AI system could augment human speech production.

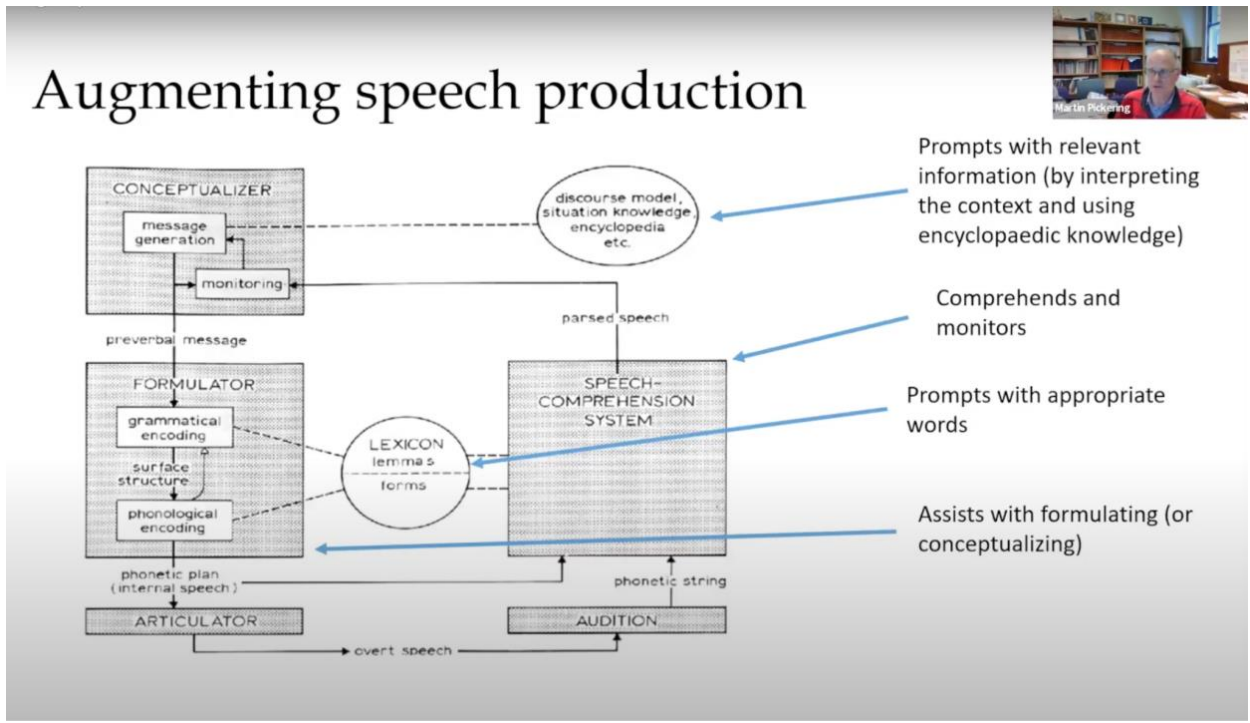

**Kyle Mahowald**'s talk distinguished between formal competence (knowledge of linguistic rules) and functional competence (real-world language use) in LLMs. He argued that while LLMs have made significant progress in formal linguistic competence, their functional competence remains uneven. Mahowald proposed using LLMs as tools for linguistic theorizing, particularly in understanding how humans generate text and produce language. Regarding human-AI collaboration, Mahowald explored the concept of "bibliotechnism," viewing LLMs as cultural technologies without beliefs of their own.

**Pablo Contreras Kallens**'s research focused on the effects of Reinforcement Learning from Human Feedback (RLHF)[3] on LLM behavior in psycholinguistic experiments. His findings suggest that RLHF fine-tuning significantly alters LLM behavior, making it more aligned with human responses (see Kristensen-McLachlan et al., in press, for discussion). This alignment includes capturing sample-level variance and better approximating human performance patterns, even in areas where humans struggle. Contreras Kallens proposed interpreting LLMs as models of intersubjective cultural patterns rather than individual language users (see Contreras Kallens & Christiansen, in press, for discussion).

**Morten H. Christiansen** discussed the impact of LLMs on textual culture, particularly their ability to generate poetry and implications for literacy. His research involved a Poetry Turing Test, as illustrated in Figure 2. The test revealed that while LLMs can mimic poetic style effectively, experts sometimes struggle to distinguish their output from human-authored poems. This work underscores LLMs' potential in creative tasks, but it also raises concerns about their impact on cognitive and cultural aspects of literacy. Christiansen highlighted opportunities, such as facilitating writing processes and creating educational materials, and pitfalls, including potential declines in cognitive skills and cultural knowledge (see Huettig & Christiansen, 2024, for discussion).

---

[3] A machine learning technique where AI models are fine-tuned based on human preferences and evaluations of their outputs (Chaudhari et al., 2024).

**Figure 2**.

An example of Christiansen's Poetry Turing Test.

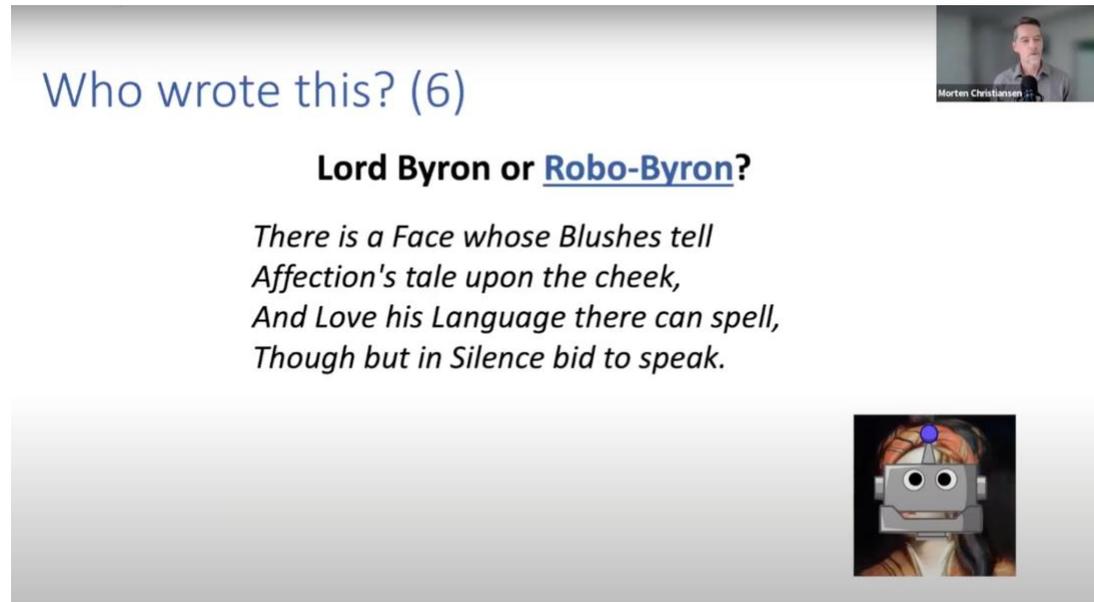

A central theme emerging from these presentations is the importance of alignment between human and LLM representations and behaviors. Pickering's concept of augmented language production and Contreras Kallens's research on RLHF provide complementary perspectives on how this alignment might be achieved and its potential benefits. As humans interact with aligned LLMs, they may experience a more natural and intuitive collaborative process, potentially leading to enhanced performance on language tasks.

While Pickering, Contreras Kallens, and Christiansen focus on the benefits and implications of human-LLM alignment and collaboration, Mahowald's presentation introduces important caveats. His distinction between formal and functional competence highlights the need for humans to navigate a complex cognitive landscape when collaborating with AI. Mahowald's exploration of bibliotechnism further complicates the picture of human-AI interaction, suggesting a need to reconsider how we conceptualize LLMs in collaborative contexts.

The integration of these viewpoints suggests that human-LLM collaboration in language tasks may result in a dynamic cognitive process that combines enhanced linguistic capabilities with critical evaluation and contextual integration. As humans become accustomed to working with LLMs, they may develop new cognitive strategies for leveraging AI capabilities while compensating for AI limitations, which could lead to a novel form of augmented cognition in language tasks.

However, the presentations also highlight the need for caution. The disparity between LLMs' formal and functional competence could lead to situations where humans must critically evaluate and contextualize LLM outputs. In these cases, humans may need to engage higher-order cognitive processes to integrate AI assistance effectively. Christiansen's research raises important questions about the impact of LLMs on literacy and cultural knowledge by emphasizing the need for evidence-based guidelines to harness LLMs positively in education while preserving the integrity of human cognitive skills.

These presentations collectively suggest that human-LLM collaboration has the potential to enhance language production, comprehension, and creativity, but it also introduces new cognitive challenges and cultural implications. As this technology continues to evolve, further research will be necessary to understand fully the implications of these human-AI interactions on cognitive processes, creativity, and literacy. Developing strategies for optimal collaboration that leverage the strengths of both human cognition and AI capabilities while addressing potential pitfalls will be essential for the future of textual culture and education in the age of LLMs.

**5.3 Topic 3: What do we know about the efficacy (potential and observed) of LLM-based tools to human learning in language-intense domains (e.g., written composition instruction)?**

The integration of LLMs into educational contexts, particularly in language-intensive domains, has sparked significant interest and debate. This section explores the potential and observed efficacy of LLM-based tools in enhancing human learning, with a focus on written composition instruction. Drawing from the presentations of **Alina Reznitskaya** and **Evgeny Chukharev** and of **Bill Cope, Mary Kalantzis,** and **Patrick Bolger**, we examine the theoretical foundations, practical applications, and implications of using AI in language education.

**Alina Reznitskaya** and **Evgeny Chukharev**'s presentation centered on the role of LLMs and assessment frameworks in promoting argumentation skills. They highlighted the limitations of traditional assessment methods, such as Toulmin's Argument Pattern (TAP), and proposed leveraging LLMs to implement more comprehensive frameworks like the Rational Force Model (RFM) and the Argumentation Rating Tool (ART). These frameworks offer a nuanced evaluation of argument quality, addressing aspects such as diversity of perspectives, clarity, acceptability of propositions, and logical validity. The presenters suggested that the integration of LLMs with symbolic AI could provide unbiased, meaningful, and specific feedback to learners, with the goal of advancing the teaching and learning of argumentation.

**Bill Cope**, **Mary Kalantzis**, and **Patrick Bolger** presented both theoretical perspectives and practical applications of generative AI in education. They posited a fundamental difference between how LLMs and humans construct meaning, with AI operating statistically and empirically, while humans use grammar theoretically. Despite this distinction, their practical application of AI in their graduate program demonstrated promising results. They reported using RAG-calibrated[4] AI in combination with human peer review to provide feedback on student projects, with that combination resulting in more extensive and finely targeted formative suggestions than would be feasible for human instructors or peer reviewers alone. The workflow for integrating the AI and peer feedback is shown in Figure 3. The presenters also discussed future plans to integrate AI into the writing process itself, with safeguards to ensure student authorship.

---

[4] Retrieval-augmented generation (RAG) is a technique that combines LLMs' generative language capabilities with information retrieval from external data sources or customized datasets. See, for example, Rangan and Yin (2024). This approach can improve the quality of AI-generated output by reducing the risk of hallucinations and providing more relevant context than using an LLM alone.

**Figure 3.**

Cope et al.'s workflow for integrating AI in combination with human peer review.

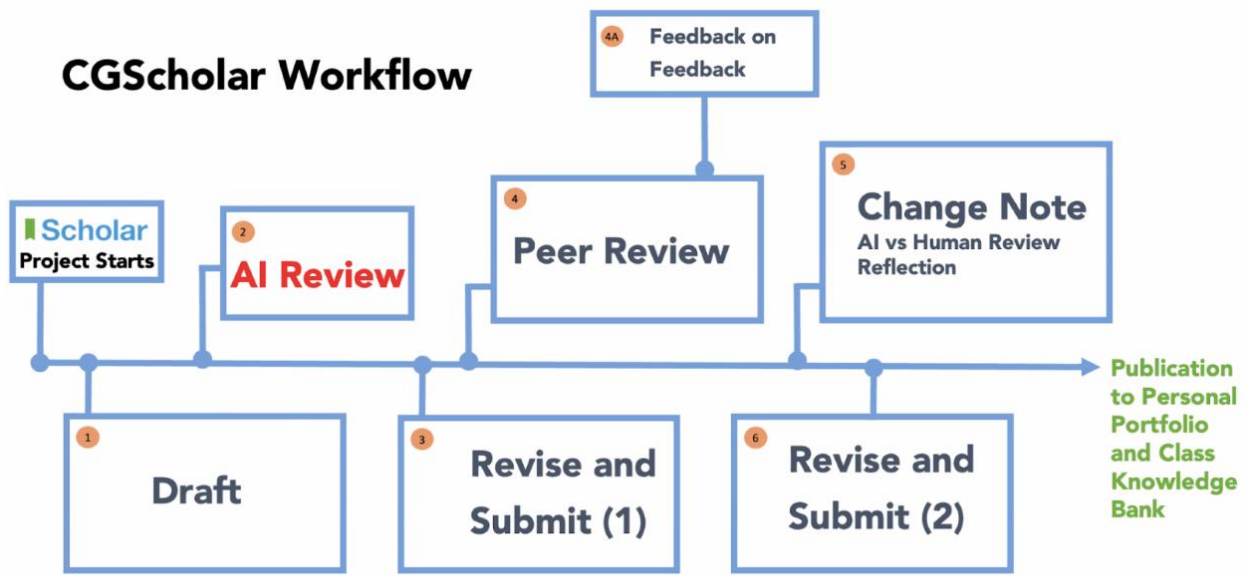

A common theme emerging from these presentations is the potential of LLM-based tools to enhance assessment and feedback in language-intensive domains. Reznitskaya and Chukharev's proposal for using LLMs to implement more sophisticated argumentation assessment frameworks aligns with Cope, Kalantzis, and Bolger's practical application of AI for providing detailed feedback on student work. Both approaches suggest that LLMs can offer more comprehensive and nuanced evaluation than traditional methods.

The presentations also highlight important considerations and challenges. Reznitskaya and Chukharev emphasized the need to ensure unbiased and meaningful feedback, proposing the integration of symbolic AI with LLMs. Similarly, Cope, Kalantzis, and Bolger discussed the importance of implementing safeguards to track student authorship and ensure AI serves as an assistive tool rather than a replacement for student effort.

In summary, these presentations collectively suggest that LLM-based tools have significant potential to enhance human learning in language-intensive domains, particularly in areas such as written composition and argumentation. The ability of these tools to provide detailed, timely feedback could lead to more personalized and effective learning experiences. Moreover, the application of AI across different educational levels, as suggested by Cope et al., indicates a wide-ranging potential impact on language education. However, realizing this potential requires careful consideration of the theoretical underpinnings of language and cognition, as well as the development of appropriate safeguards and implementation strategies. As this technology continues to evolve, further research will be necessary to fully understand its impact on learning outcomes and to develop best practices for its integration into educational contexts.

## 6. Discussion and Recommendations

The workshop's exploration of the relationship between LLMs and human cognition in text production and comprehension has yielded valuable insights into our two central research questions. While the workshop did not aim to find definitive answers, the presentations and discussions have identified promising avenues for further investigation and highlighted important considerations for the field.

Regarding our first question—to what extent LLMs offer hypotheses about how human minds produce natural-language texts—the workshop revealed a nuanced perspective. Michael Ramscar's presentation on discriminative learning and the statistical structure of language suggests that LLMs may indeed provide valuable insights into certain aspects of human language processing. His framework, which aligns more closely with how LLMs learn, proposes a shift from traditional assumptions about compositional meaning transfer to a model based on uncertainty reduction and the management of linguistic unpredictability. This alignment between LLM learning mechanisms and Ramscar's theory of human language processing indicates that these AI models could offer hypotheses about learning mechanisms in human cognition.

However, Kyle Mahowald's distinction between formal and functional competence in LLMs serves as an important caveat. While LLMs have made significant strides in mastering formal linguistic rules, their functional competence—the ability to use language effectively in real-world contexts—remains uneven. This disparity suggests that while LLMs can provide hypotheses about certain aspects of human language production, they may not fully capture the nuanced, context-dependent nature of human communication. Therefore, there may be a need for AI research to focus not just on linguistic rule-learning but also on developing models that can better navigate real-world language use contexts.

Leila Wehbe's innovative approach to using language models for studying the neurobiology of language further illustrates the potential of LLMs to inform our understanding of human language processing. Her work on fine-tuning language models to predict brain imaging data demonstrates a promising method for exploring the neural basis of language comprehension and production. This research suggests that LLMs, when used creatively, can serve as valuable tools for generating and testing hypotheses about human language processing at a neurobiological level. This approach could lead to more flexible models of brain activity during language tasks and potentially bridge the gap between language comprehension and production research.

Turning to our second question—what cognitive processes occur in human writers' minds when they collaborate with AI systems in text production—the workshop also offered intriguing perspectives. Martin Pickering's concept of augmented language production provides a framework for understanding how humans might interact with AI assistants during speech and writing tasks. His proposal that successful collaboration depends on the alignment of representations between humans and LLMs echoes findings in human-to-human dialogue studies. This suggests that as humans engage with LLMs, they may develop new cognitive strategies for leveraging AI capabilities while compensating for AI limitations, potentially leading to a novel form of augmented cognition in language tasks.

Pablo Contreras Kallens's research on the effects of RLHF on LLM behavior offers further insights into the cognitive processes at play during human-AI collaboration. His findings indicate that RLHF fine-tuning significantly alters LLM behavior, making it more aligned with human responses. This alignment extends to capturing sample-level variance and better approximating human performance patterns, even in areas where humans typically struggle. These results suggest that as LLMs become more human-like in their responses, the cognitive processes involved in human-AI collaboration may increasingly resemble those in human-human collaboration, while still requiring unique strategies for effective interaction with AI systems.

Furthermore, the workshop demonstrated that the integration of LLMs in language instruction shows promising potential, particularly in areas such as writing feedback and argumentation assessment. The work presented by Bill Cope, Mary Kalantzis, and Patrick Bolger on the practical application of AI in education provides valuable insights into the cognitive processes involved in human-AI collaboration in real-world settings. Their use of RAG-calibrated AI in combination with human peer review for providing feedback on student projects demonstrates how AI can augment human cognitive processes in educational contexts. Through a carefully designed approach, AI can provide more extensive and targeted formative feedback than traditional methods alone. This approach suggests that the integration of AI in text production tasks may lead to more extensive and finely targeted formative processes, potentially enhancing human cognitive abilities in writing and critical analysis. Similarly, Reznitskaya and Chukharev's approach to using LLMs to implement sophisticated argumentation assessment frameworks could significantly enhance writing and argumentation instruction.

However, Morten H. Christiansen's research on the impact of LLMs on textual culture raises important questions about the long-term effects of human-AI collaboration on cognitive processes. He noted potential risks such as an over-reliance on AI assistance, which could potentially hinder the development of critical thinking skills. While LLMs show remarkable capabilities in mimicking human-like text production, including poetry, for example, the implications for human creativity and cultural knowledge transmission remain unclear. This underscores the need for careful consideration of how prolonged interaction with AI systems might affect human language abilities and cognitive skills. Mitigation strategies should focus on designing AI tools that complement rather than replace human instruction, and on developing students' skills in critically evaluating AI-generated content.

**6.1 Recommendations for Ethical Use and Development of Generative AI**

Based on these insights, we propose recommendations for ethical use and development of generative AI systems. As LLMs become increasingly sophisticated, questions arise about the potential misuse of these technologies, the risk of perpetuating biases, and the importance of maintaining transparency in AI decision-making processes.

First, we recommend developing comprehensive guidelines for the ethical implementation of AI language models in educational and professional settings. These guidelines should address issues of transparency, bias mitigation, and the preservation of human agency in collaborative tasks. They should also provide a framework for assessing the impact of AI language models on human cognitive development, particularly in educational contexts. Specifically, we suggest the following actions:

1. Implement rigorous testing and ongoing monitoring to identify and mitigate biases in AI systems. This involves diverse representation in training data and regular audits of AI outputs for fairness across different demographic groups.
2. Design AI systems as collaborative tools that enhance human capabilities rather than replace them. This includes allowing users to override AI suggestions and providing clear explanations of AI-generated outputs.
3. Implement robust data protection measures, including advanced encryption, anonymization techniques, and clear consent procedures for data collection and use in AI systems.

Furthermore, we recommend establishing independent ethical review boards to assess the potential impacts and ethical implications of new AI technologies before deployment. These boards should include experts in human cognition and language processing to ensure that AI systems are developed in ways that complement and enhance human abilities rather than potentially diminishing them. This interdisciplinary approach is also helpful for ensuring diverse perspectives in AI development, minimizing potential blind spots, and addressing ethical concerns comprehensively.

**6.2 Recommendations for Future Research**

The convergence of AI language models and human cognitive processes in text comprehension and composition presents a unique challenge that requires a multifaceted approach. Drawing from the workshop discussions and identified research priorities, we propose several key recommendations to address the knowledge gap and optimize human-AI collaboration in textual tasks.

Foremost among these recommendations is the development of interdisciplinary research programs that bring together experts from cognitive psychology, linguistics, and AI, as done in this workshop. These collaborative initiatives should aim to create integrated models of human-AI language processing that account for both the statistical learning capabilities of large language models and the nuanced, context-dependent nature of human cognition. The work of Martin Pickering and Pablo Contreras Kallens has demonstrated that understanding the alignment between human and AI language processing is necessary for effective collaboration. By fostering this interdisciplinary approach through joint research projects, additional cross-disciplinary workshops, and the establishment of dedicated research centers, we can develop a more comprehensive understanding of how AI models and human minds process language, both independently and in collaboration.

Equally important is the design and implementation of targeted studies that specifically examine how human cognitive processes adapt to and interact with AI language models during collaborative text production and comprehension tasks. These studies should employ a variety of methodologies, including real-time neuroimaging, eye-tracking, and think-aloud protocols, to capture the dynamic interplay between human and AI cognition. As proposed during the workshop, comparative studies of human-human versus human-AI collaboration in various language tasks could provide valuable insights into the unique characteristics of human-AI interaction. This approach aligns with Mahowald's discussion on the differences between formal and functional competence in LLMs, potentially revealing important distinctions in how humans and AI systems approach language tasks.

The potential long-term effects of AI collaboration on human cognitive skills and literacy emerged as a critical research priority during the workshop discussions. These discussions underscored the importance of understanding how prolonged interaction with AI systems might affect human language abilities, creativity, and critical thinking skills. While Christiansen noted that there are concerns about potential negative effects such as decreased perceptual and cognitive skills, poorer domain-specific knowledge, and increased susceptibility to misinformation, there are also significant opportunities when it comes to long-term literacy impacts (Huettig & Christiansen, 2024). These include facilitating the writing process, leveling the playing field for disadvantaged writers, and generating new reading materials in under-resourced languages. These dual possibilities emphasize the critical need for careful, evidence-based approaches to AI integration in education. To address this, we advocate for the initiation of long-term longitudinal studies to assess the impact of sustained human-AI collaboration on these cognitive abilities. These studies should track individuals and groups over extended periods as they engage with AI language models in various capacities, and they should monitor changes in cognitive processes, language use patterns, and problem-solving strategies. Our recommendations for long-term studies and targeted educational interventions are designed to maximize the benefits of AI collaboration while mitigating potential risks, ensuring that AI enhances rather than diminishes human cognitive abilities and cultural knowledge.

In the educational domain, we recommend the creation and implementation of AI-enhanced interventions that leverage human-AI collaboration while explicitly addressing potential risks. Building on the work presented by Cope, Kalantzis, and Bolger on AI-assisted feedback in education, these interventions should focus on developing students' critical thinking skills in the context of AI collaboration. They should teach students to effectively evaluate and integrate AI-generated content by fostering an understanding of both the capabilities and limitations of AI language models. Future research should investigate how LLMs can be tailored to individual learner needs, learning styles, and proficiency levels to create more effective and engaging language learning environments.

Lastly, we recommend establishing comprehensive guidelines for the ethical development and implementation of AI language models that are aligned with human cognitive processes. These guidelines should address issues such as transparency in AI-generated content, the prevention of bias amplification, and the preservation of human agency in human-AI collaborative tasks. They should also provide a framework for assessing the impact of AI language models on human cognitive development, particularly in educational and professional contexts. The development of these guidelines should involve a diverse group of stakeholders to ensure a holistic approach that considers multiple perspectives and potential consequences.

By implementing these recommendations, we can work toward bridging the knowledge gap between AI capabilities and human cognitive processes in text comprehension and composition. This approach will not only advance our scientific understanding but also pave the way for more effective, ethical, and beneficial integration of AI language models in educational, professional, and everyday contexts. As we move forward, it is important that we maintain a balance between leveraging the powerful capabilities of AI and preserving the unique aspects of human cognition that drive creativity, critical thinking, and cultural knowledge transmission. The resulting knowledge will be invaluable for developing more effective AI-assisted language learning tools and ensuring that human-AI collaboration enhances rather than diminishes human language abilities.

### 7. Conclusion

The workshop fostered essential dialogue between cognitive psychology, linguistics, and AI-based NLP experts, advancing knowledge in these fields. It identified challenges, explored solutions, and created new cross-disciplinary collaboration opportunities. This interdisciplinary collaboration facilitated knowledge exchange and deepened our understanding of text comprehension and composition. The insights gained are likely to lead to more effective educational interventions, innovative learning experiences, and ethical AI applications, ultimately shaping the future of AI-human collaboration in written text processing.

**7.1 Recap of Key Insights**

The NSF-funded workshop "Text Production and Comprehension by Human and Artificial Intelligence" has yielded several significant findings that shed light on the complex relationship between artificial intelligence and human cognition. Key insights include the following:

1. LLMs offer valuable hypotheses about certain aspects of human language processing, particularly in terms of learning mechanisms and managing linguistic uncertainty.
2. Human-AI collaboration in language tasks shows promise for enhancing performance and creativity, but also introduces new cognitive challenges.
3. LLM-based tools demonstrate significant potential for improving language instruction and assessment, especially in areas such as writing feedback and argumentation analysis.
4. The integration of insights from cognitive psychology and linguistics can lead to more sophisticated and human-like AI systems, while also informing our understanding of human language processing.

**7.2 The Future of Human-AI Collaboration in Text Production**

The findings from this workshop suggest a future where human-AI collaboration in text production becomes increasingly prevalent and sophisticated, with far-reaching implications across multiple domains. Cognitively, frequent interaction with AI systems for textual tasks may lead to the development of new strategies for leveraging AI capabilities while maintaining critical thinking skills. This evolution in cognitive processes could fundamentally alter our approaches to problem-solving and information processing.

AI assistance in education could lead to more personalized and effective educational experiences, as demonstrated by Cope et al. and Reznitskaya and Chukharev. These approaches could potentially transform traditional approaches to instruction and assessment, especially in language-focused courses. However, this potential for enhancement comes with significant challenges and ethical considerations. A pressing concern is that AI could exacerbate existing educational inequalities if access is not equitable, potentially widening the gap between those who have access to advanced AI tools and those who do not. Furthermore, the use of AI in making educational decisions about students raises complex ethical questions about fairness, transparency, and the potential for bias in algorithmic decision-making.

The need for official guidelines on the appropriate use of AI in high-stakes assessments and academic writing is becoming increasingly apparent. Questions about whether AI assistance should be cited, acknowledged, or perhaps even prohibited in certain contexts need to be addressed to maintain academic integrity and ensure fair evaluation of students' abilities. These guidelines will need to evolve alongside the technology, balancing the benefits of AI assistance with the core educational objectives of developing independent thinking and skills.

On a broader societal level, the widespread use of AI in text production may profoundly influence how we communicate, create, and consume written content. This shift raises important questions about authorship, creativity, and the nature of human expression in an AI-augmented world. As AI becomes more integrated into our daily communication and creative processes, we may need to reconsider our understanding of originality and the value we place on purely human-generated content.

The potential psychological impacts of heavy reliance on AI in text production tasks, particularly for developing minds, is also an area that requires careful study and consideration. There are concerns about how constant interaction with AI language models might affect language acquisition, cognitive development, and social-emotional skills in children and young adults. Understanding and mitigating any negative effects while harnessing the positive potential of AI will be a major challenge in shaping educational policies and practices.

### 7.3 Call to Action

The rapidly evolving landscape of AI and its intersection with human cognition demands continued interdisciplinary research and collaboration. We call upon researchers, educators, policymakers, and industry leaders to:

1. Foster interdisciplinary partnerships that bridge the gaps between cognitive psychology, linguistics, computer science, and education.
2. Invest in long-term studies to understand the impact of AI collaboration on human language skills and cognitive development.
3. Develop ethical guidelines and best practices for the integration of AI in educational and professional settings.
4. Prioritize the development of AI systems that enhance human capabilities rather than replacing them.

As we advance these interrelated fields, it will be increasingly important to prioritize ethical considerations at every stage of research and development of human-AI systems and processes. This includes ensuring transparency in AI decision-making processes, protecting user privacy and data, and mitigating potential biases in AI systems.

**7.4 Closing Thoughts**

The intersection of human cognition and artificial intelligence in language processing represents one of the most transformative research areas of our time. It offers the potential to revolutionize our understanding of human language, enhance educational practices, and create powerful new tools for communication and creativity.

As we look to the future, we envision a world where human intelligence and artificial intelligence complement each other, with the potential to lead to unprecedented advancements in language processing and learning. These advancements could unlock new frontiers of human expression and understanding by pushing the boundaries of what we thought possible in language use and comprehension. Realizing this vision requires careful navigation of the challenges and ethical considerations that arise from this technological revolution. By fostering interdisciplinary collaboration, maintaining a focus on human-centered design, and prioritizing ethical development and deployment of AI technologies, we can harness the full potential of this field to transform text production and comprehension. As we continue to explore the opportunities for collaboration between human cognition and artificial intelligence, we can realize new potential for understanding and enhancing the fundamental human abilities of creating and interpreting written language. This journey may not only revolutionize how we teach and learn language skills but also deepen our understanding of the cognitive processes that underlie our ability to communicate through text.

**Funding Acknowledgement**

This material is based upon work supported by the National Science Foundation Programs in Science of Learning and Augmented Intelligence, Linguistics, and Robust Intelligence under Grant No. 2422404 (PI Evgeny Chukharev, Co-PI Emily Dux Speltz).